\documentclass{article}

\usepackage{microtype}
\usepackage{graphicx}
\usepackage{subfigure}
\usepackage{booktabs} % for professional tables

\usepackage{amsmath,amsthm,amssymb}
\usepackage{enumitem}

\usepackage{listings}
\usepackage{color}
\definecolor{mygreen}{rgb}{0,0.6,0}
\definecolor{mygray}{rgb}{0.5,0.5,0.5}
\definecolor{mymauve}{rgb}{0.58,0,0.82}

\newtheorem{definition}{Definition}
\newtheorem{proposition}{Proposition}
\newtheorem{assumption}{Assumption}
\newtheorem{remark}{Remark}

\usepackage{hyperref}

\usepackage[accepted]{icml2018}

\newcommand{\del}{\partial}
\newcommand{\R}{\mathbb{R}}

\renewcommand{\L}{\mathcal{L}}
\newcommand{\dual}[2]{\left\langle {#1}, {#2} \right\rangle}
\newcommand{\red}[1]{\textcolor{red}{#1}}

\newcommand{\eat}[1]{}
\newcommand{\mypar}[1]{\vspace{2pt} \noindent \textbf{#1}}

\icmltitlerunning{Learning in Integer Latent Variable Models}

\begin{document}

\twocolumn[
\icmltitle{Learning in Integer Latent Variable Models \\ with Nested Automatic Differentiation}





\begin{icmlauthorlist}
\icmlauthor{Daniel Sheldon}{umass,mhc}
\icmlauthor{Kevin Winner}{umass}
\icmlauthor{Debora Sujono}{umass}
\end{icmlauthorlist}

\icmlaffiliation{umass}{College of Information and Computer Sciences, University of Massachusetts Amherst}
\icmlaffiliation{mhc}{Department of Computer Science, Mount Holyoke College}

\icmlcorrespondingauthor{Daniel Sheldon}{sheldon@cs.umass.edu}

\icmlkeywords{Automatic Differentiation, Probability Generating Functions, HMM, Forward Algorithm}

\vskip 0.3in
]



\printAffiliationsAndNotice{}  % leave blank if no need to mention equal contribution

\begin{abstract}
We develop nested automatic differentiation (AD) algorithms for exact inference and learning in integer latent variable models. Recently, Winner, Sujono, and Sheldon showed how to reduce
% developed the first exact inference algorithms for a class of integer latent variable models. Their approach reduces
marginalization in a class of integer latent variable models to evaluating a probability generating function which contains many levels of nested high-order derivatives. We contribute faster and more stable AD algorithms for this challenging problem and a novel algorithm to compute exact gradients for learning. 
%Our AD approach is the first one that is polynomial instead of exponential in the number of levels of nesting.
%We also contribute a novel algorithm to compute exact gradients for learning.
These contributions lead to significantly faster and more accurate learning algorithms, and are the first AD algorithms whose running time is polynomial in the number of levels of nesting.
\end{abstract}

\section{Introduction}
\label{sec:intro}

In a recent line of work, \citet{Winner2016} and \citet{Winner2017} developed the first exact inference algorithms for a class of hidden Markov models (HMMs) with integer latent variables. Such models are used to model populations that change over time in ecology or epidemiology~\cite{Dail2011,heathcote1965}. Standard inference techniques do not apply to these models because marginalization would require summing over the infinite support of the latent variables. Instead, they showed how to reformulate the forward algorithm for HMMs to use \emph{probability generating functions} (PGFs) as its internal representation of probability distributions and messages. These PGFs can be represented compactly
%---either symbolically~\cite{Winner2016} or as a computation graph~\cite{Winner2017}---
and encode all the information about the infinite-support distributions. Inference tasks such as computing the likelihood are reduced to evaluating PGFs and their derivatives.

\begin{figure}
  \centering
  \includegraphics[width=0.8\columnwidth]{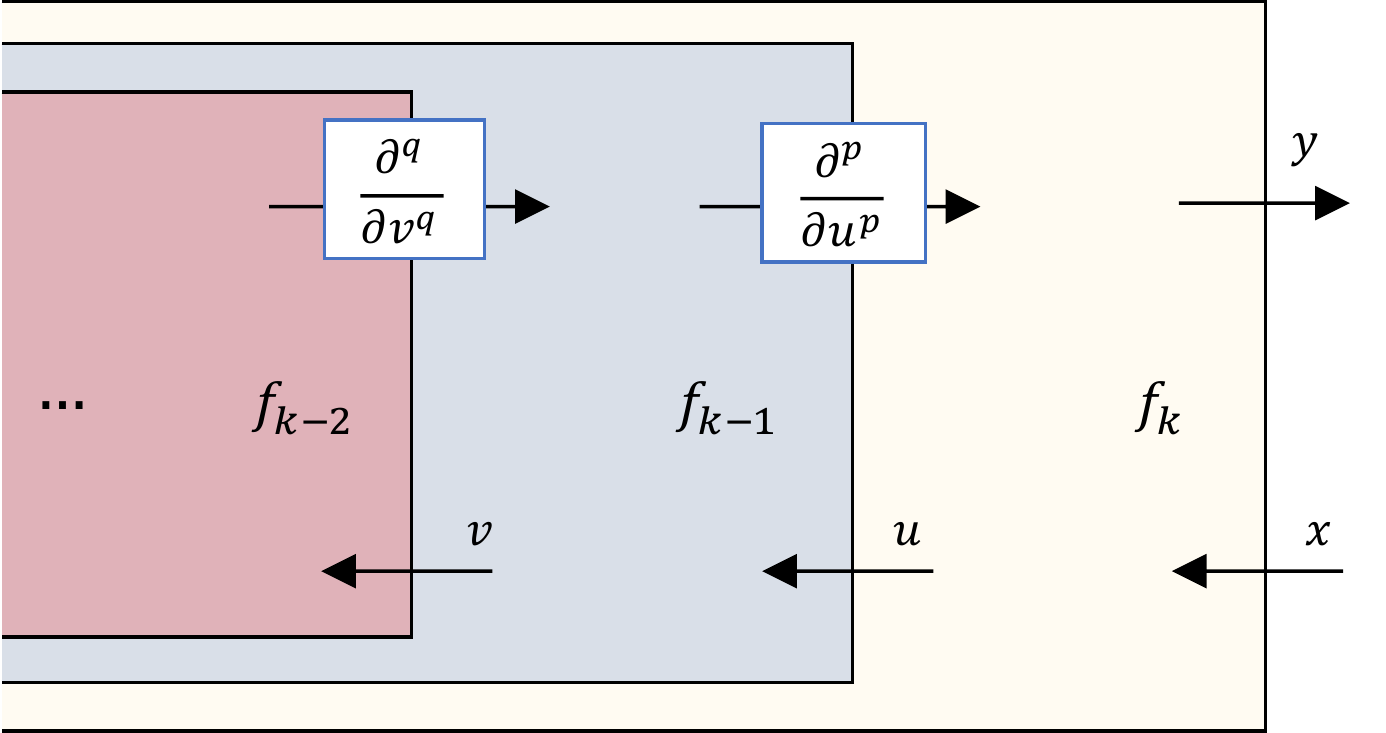}
  \caption{\label{fig:circuit}Function with nested high-order derivatives.}
\end{figure}

However, the PGFs are complex functions that are defined recursively in terms of high-order derivatives of other PGFs. Figure~\ref{fig:circuit} illustrates such a function. The function $f_k$, on input $x$, first computes a value $u$, then computes the $p$th derivative of $f_{k-1}$ with respect to $u$, and then uses the result to compute its own output value $y$. The function $f_{k-1}$ itself involves nested derivatives of $f_{k-2}$, and so on, with $k$ total levels of nesting. The goal is to compute the function $f_k(x)$, and possibly derivatives thereof. It is natural to consider automatic differentiation (AD) techniques for this task. However, this is a very difficult setting for AD. There are many levels of nesting and the total order of differentation may number in the hundreds or thousands. %\todo{if time}
%Few existing systems can handle nested derivatives, and those that do take time that scales exponentially in the number of levels of nesting.
%~\red{cites}

In this paper we subtantially improve AD algorithms for inference in integer latent variable models, and develop new AD algorithms to support learning.
For inference, we conceptually simplify the AD techniques of~\cite{Winner2017} and root them more firmly in the AD literature. We show that nested derivatives of univariate functions can be handled by an extension to the basic AD computation model that allows \emph{nested derivative nodes}, which are just derivatives of another function defined by a computation graph (e.g., the ``node'' $\frac{\del^p}{\del u^p}f_{k-1}(u)$ in Figure~\ref{fig:circuit}). The AD algorithm needs only a thin adapter to handle the change of scope between the inner and outer functions. For the inference application, this places nearly all of the complexity in the general-purpose AD toolkit and simplifies the application. 
%. For example, it allowed us to eliminate unneeded operations from the algorithms of~\citet{Winner2017}. 

We also make substantial stability and speed improvements to the AD algorithms. We show that implementing core AD operations using the \emph{logarithmic number system} (LNS) to accurately represent signed real numbers with high dynamic range allows the algorithms to scale to very high order derivatives---something which is not possible with a standard floating point representation. We also adopt fast power series composition for core AD operations~\cite{brent1978fast}, which are asympotically faster than those used in~\cite{Winner2017}.

Finally, we contribute new AD algorithms to compute exact \emph{gradients} of the log-likelihood in integer HMMs by extending (higher-order) forward-over-reverse AD to handle nested derivatives. This allows us to compute gradients $\frac{\del}{\del \theta}f(x, \theta)$ of functions like the one in Figure~\ref{fig:circuit}, where $\theta$ is a vector of parameters for the entire nested computation. We show experimentally that our new LNS-based AD algorithms are the only inference algorithms for integer latent variable models that are simultaneously accurate, fast, and stable, and that our novel algorithms for computing exact gradients lead to significantly faster and more accurate parameter estimation.

\section{Model and Problem Statement}
\label{sec:model}

%\paragraph{Integer HMM.}
We consider the \emph{integer hidden Markov model} (HMM) from~\cite{Winner2017}, an HMM with integer latent variables $n_1, \ldots, n_K$ representing a population that changes over time through the processes of immigration, reproduction, and mortality, and which is partially observed at each time step. The model is:
\vspace{-3pt}
\begin{align}
  \label{eq:latent-process}
  n_{k} &= \sum_{i=1}^{n_{k-1}}z_{k,i} + m_k \\
  \label{eq:observation-process}
  y_k  &\sim \text{Binomial}(n_k, \rho_k)
\vspace{-3pt}
\end{align}
with the initial condition $n_0 = 0$. The variable $n_k$ is the population size at the $k$th time step, and $y_k$ is the observed number, assuming that each indvidual is observed with probability $\rho_k$. The population size $n_k$ depends on an offspring process and an immigration process. First, each individual present at time $k-1$ contributes $z_{k,i}$ individuals to the present time step, where $\{z_{k, i}\}$ are iid random variables from the \emph{offspring distribution}. The ``offspring'' of one individual can include (or not include) itself, immediate offspring, or descendants of more than one generation, depending on the choices of the modeler. In particular, this distribution is used to model both survival and reproduction. Additionally, $m_k$ individuals enter the population, where $m_k$ is drawn from the \emph{immigration distribution}. The offspring and immigration distributions can be arbitrary count-valued distributions, and will be specified through their PGFs as described below.
%probability generating functions, which are described in more detail in the following section.

\paragraph{Problem Statement.}
The application goal is usually to estimate a parameter vector $\theta$ controlling the offspring and immigration distributions given some number of observations of this process~\cite{Dail2011}. We will focus on computing the log-likelihood and its gradient, which together will enable optimization routines to find maximum likelihood estimates. Let $K$ be the total number of time steps and let $y_{1:K} = (y_1, \ldots, y_K)$ (similar notation will be used throughout the paper).
%for a vector containing a contiguous range of an ordered set of variables).
Our goal is to compute $\log p(y_{1:K}; \theta)$ and $\frac{\del}{\del \theta} \log p(y_{1:K}; \theta)$.
%Other inference problems are considered in~\cite{Winner2017}, and the contributions of this paper also help solve those in a more efficient and stable way than previously possible; however, our main focus will be on learning.

\paragraph{Inference via PGFs.} Standard HMM inference algorithms such as the forward algorithm~\cite{Rabiner1989} do not apply here because the latent variables are unbounded. Hence, the messages have infinite length, and marginalization of any variable involves an infinite sum. \citet{Winner2016} and \citet{Winner2017} showed how the forward algorithm can be reformulated to use PGFs to represent messages. Standard inference tasks, such as computing the likelihood, are then converted to the problem of evaluating recursively-defined PGFs.

\begin{definition}
 The PGF of a (not necessarily normalized) probability distribution $q(n)$ is the power series $F(s) = \sum_{n=0}^\infty q(n)s^n$ with probability values as coefficients.
  %that uses the probability values as coefficients of powers of the newly introduced variable $s$. 
\end{definition}

To formulate the forward algorithm using PGFs, define $\alpha_k(n_k) := p(n_k, y_{1:k})$ and  $\gamma_k(n_k) := p(n_k, y_{1:k-1})$. These are the ``messages'' %(joint distributions of one hidden variable and  observed variables)
that are recursively computed within the standard forward algorithm. The PGFs of $\alpha_k$ and $\gamma_k$ are defined (using the corresponding capital letters) as $A_k(s_k) = \sum_{n_k=0}^\infty \alpha_k(n_k)s_k^{n_k}$ and $\Gamma_k(u_k) = \sum_{n_k=0}^\infty \gamma_k(n_k)u_k^{n_k}$. The utility of switching to a PGF representation is summarized in the following proposition.

\begin{proposition}[\citealt{Winner2016,Winner2017}]
\label{prop:PGF}
Let $F(u)$ and $G(u)$ be the PGFs for the offspring and immigration distributions, respectively.
The PGFs $\Gamma_k$ and $A_k$ satisfy the following recurrence:
\begin{align}
\label{eq:predict}
\Gamma_k(u_k) &= A_{k-1}\big(F(u_k)\big) \cdot G(u_k) \\
\label{eq:evidence}
A_k(s_k) &= \frac{(s_k \rho_k)^{y_k}}{y_k!} \cdot \Gamma_k^{(y_k)}\big(s_k(1-\rho_k)\big)
\end{align}
with the base case $A_0(s_0) = 1$. The likelihood can be recovered from the final PGFs as $p(y_{1:K}) = A_K(1)$.
\end{proposition}
Equation~\eqref{eq:predict} follows from the model definition and standard manipulations of PGFs, and is well known in the literature on branching processes~\cite{heathcote1965}. Equation~\eqref{eq:evidence} may appear surprising. It includes the $y_k$th derivative of the function $\Gamma_k$ from Equation~\eqref{eq:predict}. The derivatives are related to the selection of particular terms in the joint PGF of $n_k$ and $y_k$ corresponding to the observed value of $y_k$.

Proposition~\ref{prop:PGF} gives a recipe to compute the \emph{exact} log-likelihood and its gradient.
%, even though the latent variables are unbounded and cannot be marginalized by direct methods. To compute the log-likelihood and its gradient,
We need to compute $\log A_K(1; \theta)$ and $\frac{\del}{\del \theta} \log A_K(1; \theta)$, where $A_K$ is defined in terms of the prior PGFs through Equations~\eqref{eq:predict} and~\eqref{eq:evidence}, and we have introduced the parameter vector $\theta$, which includes the detection probabilities $\rho_k$ and any parameters of $F$ and $G$. Despite the somewhat complex appearance of the recurrence, it implies a well-defined feed-forward computation to calculate $A_K(1)$ from the constituent functions $F(\cdot)$, $G(\cdot)$ and $A_0(\cdot)$. The key complication is the fact that this function contains deeply nested high-order derivatives: $A_K$ calls $\Gamma_k^{(y_K)}$, which calls $A_{K-1}$, which calls $\Gamma_{K-1}^{(y_{K-1})}$, and so on. \citet{Winner2017} developed a method based on automatic differentation to efficiently compute $A_K(1)$, which we will extend in this paper to be more robust and efficient, and to compute gradients.

\section{Autodiff Approach}
\label{sec:autodiff}
Our problem is to compute $f(x, \theta)$ and $\frac{\del}{\del \theta} f(x, \theta)$ for the function $f(x, \theta) = \log A(x, \theta)$, which includes nested high-order derivatives. To set up the proper recursion we will generalize this to the following:
%the problem to that faced at the first level of nesting: to compute a high-order derivative of $f$, and the gradient thereof.
\vspace{-4pt}
\[
\textbf{Problem 1: }\text{compute\ \  $\frac{\del^p}{\del x^p} f(x, \theta)$\ \ and\ \  $\frac{\del}{\del \theta} \frac{\del^p}{\del x^p} f(x, \theta)$}
\vspace{-4pt}
\]
Now we will abstract away from the particular $f$ from the previous section and
consider any $f:\R^{m+1} \to \R$ defined according to a particular computation model, where $x \in \R$ will be called the \emph{input} and the vector $\theta \in \R^m$ will be called the \emph{parameters}. We will first describe how to solve both parts of Problem~1 in a basic computation model, and then extend the model to handle nested derivatives. 

%In this section, after introducing autodiff preliminaries, we will first describe how to solve the two parts of Problem 1 using forward AD and forward-over-reverse AD, respectively, for the case when $f$ \emph{does not} include nested derivatives. We will then describe how these procedures can be extended to handle nested derivatives.

\subsection{Computation Model and Dual Numbers}
%Suppose that $f$ is computed by the basic model for a feed-forward computation shown in Algorithm~\ref{alg:basic}.
The basic computation model for $f$ is shown in Algorithm~\ref{alg:basic}, following~\cite{griewank2008evaluating}.
%% \begin{enumerate}[itemsep=0pt]
%% \item Set $v_0 = x$
%% \item Set $v_j = \theta_j$ for $j = 1, \ldots m$
%% \item Set $v_j = \varphi_j(v_{A_j})$ for $j = m+1, \ldots, n$
%% \item Output $v_n$
%% \end{enumerate}
In Line 3, the function $\varphi_j$ is a \emph{primitive operation} that operates on the variables~$v_{A_j}$, where $A_j \subseteq \{0, \ldots, j-1\}$ is the set of \emph{predecessors} of $j$, and $v_{A_j} = (v_i)_{i \in A_j}$ is the subvector of $v_{0:n}$ corresponding to index set $A_j$. The predecessor relationship defines the \emph{computation graph} $G$, a directed acyclic graph (DAG) with edges from $i$ to $j$ for all $i \in A_j$. 

\mypar{Partial computations and dual numbers.}
We wish to express derivatives $\frac{dv_\ell}{dv_i}$ of a variable $v_\ell$ with respect to a preceding variable $v_i$.  To make this precise, for $i \leq \ell$, we denote the partial computation from $i$ to $\ell$ as $f_{i\to\ell}(v_{0:i})$. This is defined as the function that maps from fixed values of $v_{0:i}$ to the value of $v_\ell$ obtained by executing the procedure above starting with the assignment to $v_{i+1}$ and ending with the assignment to $v_\ell$. A formal recurrence for $f_{i \to \ell}$ is given in the supplement.
%We can now define precisely the derivative of $v_\ell$ with respect to $v_i$ using the partial computation from $v_i$ to $v_\ell$:
We can now define precisely the derivative of $v_\ell$ with respect to $v_i$:
\[
\frac{dv_\ell}{dv_i} = \frac{\del}{\del v_i} f_{i \to \ell}(v_{0:i}).
\]
\begin{definition}
  For $i \leq \ell$, a \emph{generalized dual number} $\dual{v_\ell}{dv_i}_p$ is the sequence of derivatives of $v_\ell$ with respect to $v_i$ up to order $p$:
  \vspace{-3pt}
\[
\dual{v_\ell}{dv_i}_p = \bigg(
\frac{\del^q}{\del v_i^q} f_{i \to \ell}(v_{0:i})\bigg)_{q=0}^p
  \vspace{-3pt}
\]
We say that $\dual{v_\ell}{dv_i}_p$ is a \emph{dual number of order $p$ with respect to $v_i$}. We will commonly write dual numbers as:
\vspace{-2pt}
\[
\dual{s}{du}_p = \Big( s, \frac{ds}{du}, \ldots, \frac{d^ps}{du^p}\Big)
\vspace{-2pt}
\]
in which case it is understood that $s = v_\ell$ and $u = v_i$ for some $0 \leq i \leq \ell$, and  $f_{i \to \ell}(\cdot)$ will be clear from context.
\end{definition}
This definition generalizes standard (higher-order) dual numbers by explicitly tracking the variable $v_i$ with respect to which we differentiate. In standard forward-mode autodiff, this would always be $x$. The generalization is important for nested differentation, where we will instantiate intermediate dual numbers with respect to different variables.

\begin{remark}[Dual numbers as Taylor series coefficients]
%The dual number $\dual{v_\ell}{dv_i}_p$ holds the first $p$ coefficients of the univariate Taylor expansion of $f_{i \to \ell}$ about $v_i$:
%  \[
%  f_{i \to \ell}(v_{0:i-1}, v_i + \epsilon) = \sum_{q=0}^\infty \frac{\frac{\del^q}{\del v_i^q} f_{i \to \ell}(v_{0:i})}{q!} \cdot \epsilon^q 
%  \]
%In particular,
  A dual number can be viewed as holding Taylor series coefficients. In particular, the dual number $\dual{v_\ell}{dx}_p$ encodes the first $p$ coefficients of the Taylor series of $f_{0 \to \ell}$ about $x$:
  \vspace{-3pt}
\[
f_{0 \to \ell}(x + \epsilon) = \sum_{q=0}^\infty \frac{f^{(q)}_{0 \to \ell}(x)}{q!} \epsilon^q
\vspace{-3pt}
\]
Most forward AD methods store and propagate dual numbers as truncated Taylor series~\cite{griewank2008evaluating,Pearlmutter2007}.
\end{remark}

\begin{algorithm}[tb]
   \caption{Basic computation model}
   \label{alg:basic}
   \begin{algorithmic}[1]
     \setlength{\itemsep}{1pt}
     \REQUIRE $x \in \R$, $\theta \in \R^m$, {\bfseries Output:} $f(x, \theta)$
     \STATE Set $v_0 = x$
     \STATE Set $v_{j} = \theta_j$ for $j = 1, \ldots m$
     \STATE Set $v_j = \varphi_j(v_{A_j})$ for $j = m+1, \ldots, n$
     \STATE Output $v_n$
   \end{algorithmic}
\end{algorithm}

%\subsection{Higher-Order Forward Mode and Lifting of Primitive Operations}
\subsection{Higher-Order Forward Mode}
We now discuss how to solve the first part of Problem 1---computing $\frac{\del^p}{\del x^p}f(x, \theta)$ in the basic computation model---using forward AD. Higher-order forward mode works by propagating dual numbers instead of real numbers through the computation. For this to work, each local computation must be ``lifted'' to accept dual numbers as inputs and to output a dual number. 

\begin{definition}[Lifted Function]
Let $\varphi: \R^k \to \R$ be a function of variables $u_1, \ldots, u_k$. The \emph{lifted function} $\L\varphi$ is the function that accepts as input dual numbers $\dual{u_1}{dx}_p, \ldots, \dual{u_k}{dx}_p$ and returns the dual number $\big\langle \varphi(u_1, \ldots, u_k), dx\big\rangle_p$.
\end{definition}

\begin{algorithm}[t]
   \caption{Higher-Order Forward AD}
   \label{alg:forward}
   \begin{algorithmic}[1]
     \setlength{\itemsep}{3pt}
     \REQUIRE $x, \theta$\ \ \  {\bfseries Output:} $\frac{\del^p}{\del x^p}f(x, \theta)$
     \STATE Set $\tilde{v}_0 = \dual{x}{dx}_p = (x, 1, 0, \ldots, 0)$
     \STATE Set $\tilde{v}_{j} = \theta_j$ for $j = 1, \ldots, m$
     \STATE Set $\tilde{v}_j = \L\varphi_j(\tilde{v}_{A_j})$ for $j = m+1, \ldots, n$
     \STATE Extract and return the $p$th derivative from $\tilde{v}_n$
   \end{algorithmic}
\end{algorithm}
\begin{algorithm}[t]
  \caption{Forward-Over-Reverse AD}
  \label{alg:forward-over-reverse}
  \begin{algorithmic}[1]
    \setlength{\itemsep}{3pt}
    \REQUIRE $x, \theta$\ \ \  {\bfseries Output:} $\frac{\del}{\del\theta} \frac{\del^p}{\del x^p}f(x, \theta)$
    \STATE Run forward AD (Algorithm~\ref{alg:forward}) to compute $\tilde{v}_0, \ldots, \tilde{v}_n$
    \STATE Initialize $\bar{v}_n = \dual{1}{dx}_p = (1, 0, \ldots, 0)$
    \STATE Initialize $\bar{v}_i = \dual{0}{dx}_p = (0, 0, \ldots, 0)$ for $i < n$
    \STATE For $j=n$ down to $m+1$ and for all $i \in A_j$:
    \begin{align*}
      \bar{v}_i &\leftarrow \bar{v}_i + \bar{v}_j \cdot \L \frac{\del}{\del v_i} \varphi_j\big(\tilde{v}_{A_j}\big)
    \end{align*}
    \STATE Extract and return the $p$th derivatives from $\bar{v}_1, \ldots, \bar{v}_m$.
  \end{algorithmic}
\end{algorithm}

%% \textbf{Higher-order forward mode for $\frac{\del^p}{\del x^p}f$:}
%% \begin{enumerate}[itemsep=0pt]
%% \item Set $\tilde{v}_0 = \dual{x}{dv_0}_p = (x, 1, 0, \ldots, 0)$
%% \item Set $\tilde{v}_j = \theta_j$ for $j = 1, \ldots, m$
%% \item Set $\tilde{v}_j = \L\varphi_j(\tilde{v}_{A_j})$ for $j = m+1, \ldots, n$
%% \item Output $\tilde{v}_n$.
%% \end{enumerate}

We defer the details of \emph{how} a function is lifted for the moment. If we are able to lift each primitive operation in the original procedure, we can execute it using dual numbers (with respect to $x$) in place of real numbers. The procedure is shown in Algorithm~\ref{alg:forward}.
The output $\tilde{v}_n$ is a dual number with respect to $x$, from which we can extract the derivatives $\frac{d^q v_n}{dx^q} = \frac{\del^q}{\del x^q}f(x, \theta)$ for $q=0, \ldots, p$.
%and, in particular, the desired value $\frac{\del^p}{\del x^p}f(x, \theta)$.
Note that in Line 2, we did not write the parameter values as dual numbers; we will allow in our notation a real number $\theta$ to be used as an input to a lifted operation $\L\varphi$, with the understanding that: (1)~$\theta$ does not depend on $x$, and (2)~$\theta$ will be promoted to the dual number $\dual{\theta}{dx}_p = (\theta, 0, \ldots, 0)$. 

It is well known how to lift basic mathematical operations:
\begin{proposition}[\citealt{griewank2008evaluating}]
  The arithmetic operations $x + cy$, $x * y$, $x / y$, $x^r$, viewed as functions of $x$ and $y$, and the mathematical functions $\ln(x)$, $\exp(x)$, $\sin(x)$, $\cos(x)$ can be lifted to operate on dual numbers of order $p$. In each case, the lifted function runs in $O(p^2)$ time.
  \label{prop:lift}
\vspace{-5pt}
\end{proposition}
As a result,
%for any function $f$ written as a procedure that uses only these primitive operations,
forward-mode autodiff can compute the first $p$ derivatives of any function $f$ that uses only these primitive operations in time $O(p^2)$ times the running time of $f$.

\eat{
So, a dual number is simply a data structure that holds a sequence of derivatives, and we will see that higher-order forward mode autodiff works by propagating dual numbers instead of scalar values. This is a conceptually clean way to think about it and hides the implementation details of \emph{how} these derivative sequences are propagated.
%---it is reasonable to think that the first $p$ derivatives of $f(x)$ with respect to $u$ depend only on the first $p$ derivatives of $x$ with respect to $u$ and the function $f$---
We will see that in practice, dual numbers are viewed as the coefficients of Taylor series.
}

\subsection{Forward-Over-Reverse AD}
For the purpose of learning, we need to solve the second part of Problem 1 and compute the gradient 
$
\frac{\del}{\del \theta} \frac{\del^p}{\del x^p} f(x, \theta).
$
%We first describe how to do this in the absence of nested derivatives.
%, i.e., in the case where we are given a computation graph for $f$ that does not have any nested derivative nodes.
This can be accomplished using ``forward-over-reverse'' AD, which is based on the relatively simple observation that we can switch the order of differentation to see that
\[
\frac{\del}{\del \theta} \frac{\del^p}{\del x^p} f(x, \theta)
= \frac{\del^p}{\del x^p}\frac{\del}{\del \theta} f(x, \theta),
\]
which reveals that what we want is also the higher-order derivative of $g(x,\theta) := \frac{\del}{\del \theta} f(x, \theta)$ with respect to $x$. The well-known \emph{reverse} mode of autodiff (or backpropagation) provides a procedure to compute $g(x; \theta)$. Then, by executing reverse mode using dual numbers instead of real numbers, we can obtain $\frac{\del^p}{\del x^p}g(x; \theta)$.
%This leads to what is sometimes called \emph{forward-over-reverse} autodiff~\cite{}.
%
Forward-over-reverse AD is shown in Algorithm~\ref{alg:forward-over-reverse}. It incrementally computes the \emph{adjoint}:
\[
\bar{v}_i
%:= \frac{\del}{\del v_i} \frac{\del^p}{\del v_0^p} f_{i \to n}(v_{0:i})
:= \frac{\del^p}{\del x^p} \frac{\del}{\del v_i} f_{i \to n}(v_{0:i}),
\]
for all $i$ in a reverse sweep through the computation graph.
%The adjoint is defined as:
%i.e., the (higher-order derivative with respect to $x$ of the) senstivity of the final output with respect to $v_i$ in the partial computation from $v_i$ to $v_n$.
%The reader will recognize that Algorithm~\ref{alg:forward-over-reverse} is just the standard reverse mode executed using dual numbers instead of real numbers.
The final adjoint values are dual numbers, from which the derivatives with respect to $x$ of the parameter gradients can be extracted. As in the forward mode, we need to lift the primitive operations of the procedure. In this case the primitive operations are those that compute the partial derivatives of $\varphi_j$ with respect to its own inputs. When $\varphi_j$ is a simple mathematical operation ($+$, $\times$, $/$, $\exp$, $\log$, $\sin$, etc.), these partial derivatives are also simple and can be lifted by standard techniques (cf. Proposition~\ref{prop:lift}).

\subsection{Forward Mode with Nested Derivatives}

%So far, we have not described how to handle the nested derivatives required for our application. 
We now extend the forward AD algorithms to handle nested derivatives within the computation procedure. For the most part, we use ideas that were present in~\cite{Winner2017}. However, we simplify the conceptual framework considerably, so that all we need to do is lift a function $\varphi_j$ that takes the derivative of another function, which leads to a conceptual improvement of the AD techniques and a dramatic simplification of the inference application.
%This simplification is significant for several reasons. First, it allows us to compare the technique to existing autodiff techniques. We observe that our method, adapted from~\citet{Winner2017}, cleanly handles nested derivatives of univariate functions without risk of ``perturbation confusion''~\cite{}\todo{missing citation}. Furthermore, it does so with running time that grows only \emph{polynomially} in the number of nesting levels; existing methods that can solve this problem all scale \emph{exponentially} in the number of nesting levels.
%However, the method doesn't presently handle arbitrary nested derivatives, \red{as we detail below}.
%Finally, our simplification leads to a simpler and more efficient implementation of PGF-based inference---the complexity of computing nested derivatives is now encapsulated in the AD procedure, so the application code is much simpler, and we avoid some unneeded operations from~\citet{Winner2017}.
%by cleanly encapsulating the tools needed to handle nested differentiation within the autodiff toolkit, so the application-specific implementation can be much simpler. 

\mypar{Extended computation model: nested derivative nodes.}
%We will extend the basic computation model (Algorithm~\ref{alg:basic}) to allow for nested derivatives.
Although the functions $\varphi_j$ are usually conceptualized as simple ``primitive'' operations, they may be arbitrarily complex as long as we know how to lift them---i.e., modify them to propagate dual numbers. So, assume now that one or more of the $\varphi_j$ functions is a \emph{nested derivative node}, which takes the derivative of some other function $g$ with respect to one of its inputs:
\vspace{-8pt}
\begin{equation}
  \label{eq:nested-node}
  \varphi_j(v_{A_j}) := \frac{\del^q}{\del v_k^q } g(v_k, \pi).
\vspace{-3pt}
\end{equation}
%Note that the expression in~\eqref{eq:nested-node} is in the same form as our original function: it is a higher-order derivative of a scalar-valued function with respect to one of its inputs $v_k$, and
Here we consider $\pi = v_{A_j \setminus k}$ to be ``parameters'' of the nested computation. We need to reason about how to lift $\varphi_j$ to propagate dual numbers with respect to $x$. We will make the following restriction on $\varphi_j$, with which we can reason about the sensitivity of $\varphi_j$ to $x$ through $v_k$ alone:
\begin{assumption}
There is no path in $G$ from $x$ to $\pi$.
\end{assumption}
\noindent \textbf{Lifting a univariate primitive via composition of Taylor series.} 
We wish to lift a nested derivative node $\varphi_j(v_k, \pi)$. To do so, we will discuss the general procedure to lift a univariate function $\varphi_j(v_k)$. We may temporarily suppress the parameters $\pi$ from notation because they are constant with respect to $x$. 
Here is the general setting. We have completed the partial computation to compute the first $p$ derivatives of $v_k = f_{0 \to k}(x)$  with respect to $x$. We wish to compute the first $p$ derivatives of $v_j = \varphi_j( v_k ) = \varphi_j \big( f_{0 \to k}(x) \big) = f_{0 \to j}(x)$.
This can be done by composing the Taylor series of $\varphi_j$ and $f_{0 \to k}$.
\begin{proposition}
  \label{prop:lift-taylor}
  Suppose $\varphi_j$ and $f_{0 \to k}$ are analytic.
  Let $\varphi_j(v_k + \tau) = v_j + \sum_{i=1}^\infty q_i \tau^i := v_j + Q(\tau)$ be the Taylor series expansion of $\varphi_j$ about $v_k$, and let $f_{0 \to k}(x + \epsilon) = v_k + \sum_{i=1}^\infty r_i \epsilon^i := v_k + R(\epsilon)$ be the Taylor series expansion of $f_{0 \to k}$ about $x$.
  %Let $Q(\tau) = \varphi_j(v_k + \tau) - v_j = \sum_{i=1}^\infty q_i \tau^i$ be the Taylor series expansion of $\varphi_j$ about $v_k$, minus the constant term $v_j$, and let $R(\epsilon) = f_{0 \to k}(x + \epsilon) - v_k = \sum_{i=1}^\infty r_i \epsilon^i$ be the Taylor series expansion of $f_{0 \to k}$ about $x$, minus the constant term $v_k$.
  Then the Taylor series expansion of $f_{0 \to j}$ about $x$ is:
  \vspace{-8pt}
  \[
  f_{0 \to j}(x + \epsilon) = v_j + Q(R(\epsilon)).
  \vspace{-3pt}
  \]
The first $p$ coefficients of $\epsilon$ in $Q(R(\epsilon))$ encode the first $p$ derivatives of $f_{0 \to j}$, and can be computed by power series composition algorithms from the first $p$ coefficients of the power series $Q$ and $R$.
\end{proposition}

\begin{algorithm}[t]
   \caption{$\textsc{Compose}\big(\dual{u}{dv}_p , \dual{v}{dx}_p\big)$}
   \label{alg:compose}
   \begin{algorithmic}[1]
     \setlength{\itemsep}{3pt}
     %\REQUIRE Dual numbers $\dual{u}{dv}_p$ and $\dual{v}{dx}_p$
     \STATE Unpack dual numbers to Taylor coefficients $q_i = \frac{1}{i!} \frac{d^iu}{dv^i}$ and $r_i = \frac{1}{i!}\frac{d^iv}{dx^i}$ for $i = 1$ to $p$, and scalar value $u$
     \STATE Let $Q(\tau) = \sum_{i=1}^p q_i \tau^i + O(\tau^{p+1})$
     \STATE Let $R(\epsilon) = \sum_{i=1}^p r_i \epsilon^i + O(\epsilon^{p+1})$
     \STATE Compute the first $p$ coefficients $s_1, \ldots, s_p$ of $\epsilon$ in the power series $S(\epsilon) = Q(R(\epsilon))$ using a power series composition algorithm.
     \STATE Return $\dual{u}{dx}_p = (u, 1! s_1, 2! s_2, \ldots, p! s_p)$
   \end{algorithmic}
\end{algorithm}

\begin{algorithm}[t]
   \caption{$\textsc{Diff}_q\big(\dual{y}{dv}_{q+p}\big)$} %$D^q \dual{y}{dv}_{q+p}$}
   \label{alg:diff}
   \begin{algorithmic}[1]
     \setlength{\itemsep}{3pt}
     %\REQUIRE Dual number $\dual{y}{dv}_{q+p}$ %, \ \textbf{Output: } $D^q\dual{y}{dv}_{q+p}$
     \STATE Return $\displaystyle \Big(\frac{d^q y}{dv^q}, \ldots, \frac{d^{q+p}y}{dv^{q+p}} \Big)$ \hfill // shift left $q$ positions
   \end{algorithmic}
\end{algorithm}

\begin{algorithm}[t]
  \caption{Lifted Nested Derivative $\L \varphi = \L \,\frac{\del^q}{\del v^q} g(\cdot, \pi)$}
  %\caption{$\L \frac{\del^q}{\del v_q} g\big( \dual{v}{dx}_p,  \pi \big)$ Lifted Nested Derivative }
  %\caption{$\textsc{LiftDiff}_q\big( g, \, \dual{v}{dx}_p, \, \pi \big)$}
   \label{alg:lift-nested}
   \begin{algorithmic}[1]
     \setlength{\itemsep}{3pt}
     \REQUIRE Function $g(v, \pi)$, dual number $\dual{v}{dx}_p$, and $\pi$, which does not depend on $x$
     %, function $\varphi_j(v, \pi) = \frac{\del^q}{\del v^q} g(v, \pi)$
     %; no path from $x$ to $\pi$ in $G$
     %\ENSURE Dual number $\dual{u}{dx}_p = \L \varphi_j\Big( \dual{v}{dx}_p, \, \pi \Big)$
     %\ENSURE Dual number $\dual{u}{dx}_p = \L \frac{\del^q}{\del v_q} g\Big( \dual{v}{dx}_p, \, \pi \Big)$
     %\textbf{Output:} $\dual{u}{dx}_p$
     \STATE Unpack scalar $v$ from $\dual{v}{dx}_p$ and initialize a new dual number $\dual{v}{dv}_{q+p}$ for the inner scope of $\varphi_j$
     \STATE Let $\dual{y}{dv}_{q+p} = \L g\big(\dual{v}{dv}_{q+p}, \pi \big)$ be the result of
     the lifted version of $g$ on the newly initialized dual number (recursively apply forward-mode autodiff to $g$) 
     %\STATE Let $\dual{u}{dv}_q = D^q \dual{y}{dv}_{q+p}$
     %\STATE Return \textsc$\dual{u}{dv}_p \circ \dual{v}{dx}_p$
     \STATE Let $\dual{u}{dv}_p = \textsc{Diff}_q\big(\dual{y}{dv}_{q+p}\big)$
     \STATE Return \textsc{Compose}$\big(\dual{u}{dv}_p , \dual{v}{dx}_p\big)$
   \end{algorithmic}
\end{algorithm}

Proposition~\ref{prop:lift-taylor} (proved in the supplement) is the foundation of higher-order forward mode. The input to $\L \varphi_j$ is the dual number $\dual{v_k}{dx}_p$, which gives us the first $p$ coefficients of $R(\epsilon)$, i.e., $r_i = \frac{1}{i!}\cdot \frac{d^i v_k}{dx^i}$. The function $\varphi_j$ is typically a simple mathematical primitive (e.g., $\log, \exp, \sin$), which will have a very special Taylor series $Q(\tau)$. The goal is to compute the first $p$ coefficients of $Q(R(\epsilon))$. In most cases of ``true'' primitive operations, specialized routines exist to do this in $O(p^2)$ or $O(p \log p)$ time~\cite{griewank2008evaluating,brent1978fast}. For general $\varphi_j$, \citet{brent1978fast} describe fast algorithms to compute the first $p$ coefficients of $Q(R(\epsilon))$ from the first $p$ coefficients of $Q$ and $R$. The compose operation we implement will run in $O(p^{2.5})$ time---see Section~\ref{sec:application}.
%in running time $O(p^{2.5})$ or faster.
%---in the best case in $O((p \log p)^{3/2})$ time.
%. The fastest asymptotic running time for power series composition is $O((p \log p)^{3/2})$, though a different algorithm will be more practical in our setting.
%The truncated power series composition $Q(R(\epsilon))$ can also be viewed as an application of the higher-order chain rule commonly known as Fa\`a d\'i Bruno's formula~\cite{wheeler1987bell}, though power series composition is far more practical computationally.

Algorithm~\ref{alg:compose} shows the \textsc{Compose} operation for two dual numbers $\dual{u}{dv}_p$ and $\dual{v}{dx}_p$, which extracts Taylor coefficients from the dual numbers and then performs power series composition to compute $\dual{u}{dx}_p$.

\mypar{Lifting a nested derivative node.} We now return to the specifics of lifting a nested derivative node $\varphi \big(v, \pi \big) = \frac{\del^q}{\del v^q}g(v, \pi)$ to propagate dual numbers with respect to $x$ (we temporarily drop subscripts without ambiguity). Note that the semantics of the partial differentiation operation $\frac{\del^q}{\del v^q}$ would be simple if we were propagating dual numbers with respect to $v$: we would just shift all derivatives in the sequence lower by $q$ positions (see the \textsc{Diff} operator in Algorithm~\ref{alg:diff}). So, our solution to nested derivatives will be to temporarily instantiate a new dual number \emph{with respect to $v$} within the scope of $\L \varphi$, propagate this through $\L g$ to compute the derivatives of $g$ with respect to $v$, apply the \textsc{Diff} operator to the result, and then apply the \textsc{Compose} operation to convert back to a dual number with respect to $x$. The whole procedure is detailed in Algorithm~\ref{alg:lift-nested}.

By using this procedure to lift nested derivative nodes, we can now run forward AD in the extended computation model. Furthermore, because we lift the nested function by recursively applying forward AD to it, we can handle recursively nested derivatives (of univariate functions, i.e., subject to Assumption 1). One can prove inductively that the running time
%of the entire procedure
increases by a modest factor, regardless of the number of levels of nested derivatives. Specifically, let $d$ be the maximum order of differentiation, which, for a nested derivative node, is equal to its own order \emph{plus} that of the calling procedure.
%This means that $d$ is an upper bound for the number of nesting levels.
In the lifted procedure, each operation takes $O(T(d))$ times its original running time, where $T(d)$ is the maximum time of an operation on dual numbers of order $d$, which in our models is the \textsc{Compose} operation.
%is the most expensive, so $T(d)$ is the time to compose two power series of order $d$.
%To summarize:
\vspace{-2pt}
\begin{proposition}
Higher-order forward AD can be extended to handle nested univariate derivatives
  %in our extended computation model
by recursively calling forward AD using Algorithm~\ref{alg:lift-nested}. The running time is bounded by $O(T(d))$ times the number of primitive operations, where $d$ is the maximum order of differentiation, and $T(d)$ is the time to compose two power series of order $d$.
\end{proposition}
\vspace{-2pt}

\subsection{Nested Forward-Over-Reverse}
We now wish to extend forward-over-reverse AD to the computation model in which $\varphi_j$ can be a nested derivative node. By examining Algorithm~\ref{alg:forward-over-reverse}, we see in Line 4 that we will need to lift a function of the form $\frac{\del}{\del v_i} \varphi_j \big(v_{A_j} \big) = \frac{\del}{\del v_i}\frac{\del^q}{\del v_k^q}g(v_k, \pi)$. By again switching the order of differentiation, the function to be lifted is:
\[
%\frac{\del^q}{\del v_k^q} \frac{\del}{\del v_i} g(v_k, \pi)
\frac{\del^q}{\del v_k^q} \bigg(\frac{\del}{\del v_i} g(v_k, \pi)\bigg)
\]
where $\frac{\del}{\del v_i} g(v_k, \pi)$ is the scalar-valued function that computes the partial derivative of $g(v_k, \pi)$ with respect to $v_i$. Observe that this is just another nested derivative node---for the function that computes the partial derivative of $g$
%instead of the value of $g$
---which we can lift using Algorithm~\ref{alg:lift-nested}. In practice, we don't need to apply Algorithm~\ref{alg:lift-nested} separately for each partial derivative. We can compute $\frac{\del^q}{\del v_k^q} \frac{\del}{\del v_i} g(v_k, \pi)$ for \emph{all} $i$ simultaneously by (recursively) applying forward-over-reverse AD to $g$, within a procedure similar to Algorithm~\ref{alg:lift-nested} (details omitted). This will be more efficient because it only needs one forward and reverse sweep through the computation graph for $g$. In summary:

\begin{proposition}
Forward-over-reverse AD can also be extended to handle nested derivative nodes by recursively calling forward-over-reverse AD using Algorithm~\ref{alg:lift-nested} (or a more efficient variant). The running time is $O(T(d))$ times that of the original computation.
\end{proposition}

\section{Application and Implementation}
\label{sec:application}

We now return to inference and learning in integer HMMs. The log-likelihood $f(x, \theta) = \log A_K(x; \theta)$ (Equation~\ref{eq:predict}) clearly fits within the extended computation model: it involves only basic mathematical primitives and a nested derivative of $\Gamma_K$.
%The function $\Gamma_K$ is similar (Equation~\ref{eq:evidence}): it invokes $A_{K-1}$, which includes a nested derivative of $\Gamma_{K-1}$.
Therefore, we can apply the nested forward-over-reverse AD algorithms from Section~\ref{sec:autodiff} to compute the log-likelihood and its gradient. Observe that the total number of levels of nesting is $K$ and the maximum order of differentiation is $d = \sum_k y_k$.
%, the number of time steps in the HMM, , the sum of the observed counts. 

This is a unique application of AD. 
The total order $d$ may number in the hundreds or thousands with ten or more levels of nesting. For comparison, typical values of $d$, even in high-order applications, are single digits~\cite{griewank2008evaluating}.
%Applications of nested AD are rare; reported examples involve one level of nesting~\cite{Siskind2008}, e.g., to solve bilevel optimization problems that arise in machine learning such as hyperparameter optimization~\cite{foo2008efficient,maclaurin2015gradient} or minimizing a loss function that depends on per-example optimizations to make predictions~\cite{domke2012generic}.
Applications of nested AD are rare; reported examples involve only one level of nesting~\cite{Siskind2008,foo2008efficient,maclaurin2015gradient,domke2012generic}.
Because of the very high order and nesting, we faced significant implementation challenges related to two interrelated issues: numerical stability and asymptotic efficiency of power series operations. We describe in this section the steps needed to overcome these.

\subsection{Numerical Stability: Logarithmic Number System}
The core computations in our approach are lifted primitive operations for dual numbers (Proposition~\ref{prop:lift}), and the \textsc{Compose} operation. Both operate on Taylor series coefficients of the form $r_i = \frac{1}{i!}\frac{d^iv}{dx^i}$ for $i$ from $0$ to $d$.
Due to the $\frac{1}{i!}$ factor in each coefficient and the fact that the derivatives themselves may have high dynamic range, overflow and underflow are significant problems and unavoidable for $d$ greater than a few hundred if coefficients are stored directly in floating point. Such problems are familiar in probabilistic inference, where a standard solution is to
%implement the algorithm ``in log-space'' by
log transform all values and use stable operations such as logsumexp whenever this transformation needs to be undone. 

Our solution is similar, but more difficult for several reasons. To see why, it is helpful to see the nature of algorithms to lift primitive operations, which use convolutions or related recurrences~\cite{griewank2008evaluating}.
%Power series composition algorithms use lifted addition and multiplication as subroutines.
As an example, here is the recurrence for the division operation to compute the Taylor coefficients $\{s_i\}$ of the function $1/v(x)$ given the Taylor coefficients $\{r_i\}$ of $v(x)$:
\[
s_i = \frac{1}{r_0}\Big[ r_i - \sum_{j=0}^{i-1} s_j r_{i-j}  \Big]
\]
%, and it is unclear how to avoid these problems in a floating point representation (e.g., to underflow gracefully without causing downstream errors).
Observe that we must handle subtraction and negative values, 
%recurrences may involve subtraction, and that coefficients may be negative, so we must deal with negative values,
unlike standard probabilistic inference. Also, \emph{both} multiplication and addition are required in the innermost loop of the algorithm (the right hand side includes multiplication and addition of the most recently computed value $s_{i-1}$)
%For example, the right hand side includes multiplication and addition of the most recently computed value $s_{i-1}$.
so it is not possible to batch multiplications and additions, with transformation to and from log-space only between batches.
%, as is often possible with probabilistic inference.
%for $s_i$ utilizes multiplication and addition with the most recently computed value $s_{i-1}$.

Our solution, which generalizes the ``log-space trick'' for probabilistic inference, is to store coefficients using a \emph{logarithmic number system}~(LNS,~\citealt{swartzlander1975sign}), and implement core power series operations in LNS. In LNS, a real number is represented as $X = s \cdot b^x$ where $s = \text{sign}(X)$ is the sign bit and $x = \log_b(|X|)$ is the log of the magnitude of $X$, stored in fixed precision.
%LNS has been proposed as an alternative to floating point for implementation in computer hardware, but is not widely adopted.
The multiplication, division, and power operations are simple and efficient in LNS, while addition and subtraction require more effort. For example, if $X$ and $Y$ are both positive, then $\log_b(X + Y) = x + \log_b\big(1 + b^{y-x}\big)$. We implemented LNS in C using a floating-point instead of fixed point internal representation to interface more easily with existing mathematical functions. Addition and subtraction use the C standard library's \texttt{log1p} function, and are much more expensive than floating point operations, despite the fact they are, in principle, single ``arithmetic operations'', and could be comparable to floating point operations with appropriate hardware~\cite{coleman2000}.

\subsection{Fast and Accurate Power Series Operations}
The lifting procedures for primitive operations that we describe above
%, which correspond to arithmetic operations on Taylor series,
all take $O(d^2)$ time and have recurrences that resemble convolutions.
%Power series
Multiplication can be done in $O(d \log d)$ time using FFT-based convolution, and
%, with FFT-multiplication as a subroutine,
$O(d\log d)$ algorithms for most other primitives can be also be derived using the FFT~\cite{brent1978fast}.
%\citet{griewank2008evaluating} observe that FFT-based algorithms are unlikely to be faster in practice for typical problems (e.g., $d < 20$).
Although our application certainly reaches the regime where FFT is faster than direct convolution, existing FFT implementations require that we transform coefficients out of LNS into floating point, 
%we need to transform coefficients from LNS to floating point to use existing FFT implementations, and
which led to inaccurate results and numerical problems when we tried this. Similar observations have been made in previous applications of FFT for probabilistic inference~\cite{Wilson2016}. %, which we confirm in our experiments.
%and our experiments show that the FFT also leads to poor accuracy when used in our baseline, the truncated forward algorithm.
A possible future remedy is to implement the FFT in LNS~\cite{Swartzlander1983}, however, our current implementation uses the direct $O(d^2)$ algorithms. 

The \textsc{Compose} operation is a bottleneck and should be implemented as efficiently as possible. \citet{Winner2017} used a naive $O(d^3)$  composition algorithm. \citet{brent1978fast} present two fast algorithms---BK 2.1 and BK 2.2---for power series composition. BK 2.2 has the fastest known running time of $O\big((d \log d)^{1.5} \big)$ with FFT convolution, but may be slower in practice~\cite{johansson2015fast} and we have already ruled out FFT convolution. Instead, we use BK 2.1, which, in our setting, runs in $O(d^{2.5})$ time and achieves substantial speedups over the naive $O(d^3)$ algorithm.

%% * Elemental operations: FFT poor accuracy. Cite Keich papers
%% * Use recurrences: $O(d^2)$ 
%% * Composition is a computational bottleneck. Naive algorithm used by \citet{Winner2017}: $O(d^3)$
%% * Brent-Kung Algorithm 2.2 $O\big((d \log d)^{1.5} \big)$ fasted asympotically but uses FFT multiplication --> unstable
%% * Brent-Kung Algorithm 2.1 using classical matrix multiplication algorithm: $O(d^{2.5})$.
%% * Can achieve some savings due to sharing some of this work across multiple compose operations (omit?)

\section{Experiments}
\label{sec:experiments}

\eat{
  \subsection{Nested Autodiff}
  Compare our method vs. exponential methods. E.g., HIPS autograd, tower method in Haskell.
}

We conducted experiments on the accuracy, speed, and stability of AD inference algorithms for integer HMMs, and of parameter estimation using AD gradient algorithms.
%and the speed and accuracy of 

\newcommand{\trunc}{{\textsc{Trunc}}}
\newcommand{\truncfft}{\textsc{Trunc-FFT}}
\newcommand{\ad}{\textsc{AD}}
\newcommand{\adlns}{\textsc{AD-LNS}}

\noindent
\textbf{Inference: Accuracy, Speed, Stability.}
We evaluated four methods for inference in integer HMMs. \trunc{} and \truncfft{} are variants of a \emph{truncated forward algorithm}, which is currently used in practice in ecological applications~\cite{Dail2011}. These place an \emph{a priori} upper bound $N$ on the values of the hidden variables and then apply the standard forward algorithm for discrete HMMs. They require a convolution of the offspring and immigration distributions to compute the transition probabilities (details are given in the supplement), and take $O(KN^3)$ time with direct convolution and $O(K N^2 \log N)$ time with FFT convolution. They can be implemented in log-space easily, but FFT convolutions must be done in linear space, which can lead to accuracy loss~\cite{Wilson2016}. Selecting $N$ in the truncated algorithms is a challenge. A too-small value will cut off some of the probability mass and lead to the wrong result, and a too-large value will lead to high running time.
%One way to select $N$ in practice would be to increase it until the answer stabilizes.
For these experiments, we iteratively doubled $N$ until the log-likelihood converged (to a precision of 5 decimal places) or until $N$ reached an absolute limit of 2500, and \emph{only measured the running time of the final iteration}. Note that this is a conservative comparison that hides the cost of tuning this value in practice.

The \ad{} and \adlns{} methods use nested AD to compute $\log A_K(1; \theta)$, with the latter storing coefficients in LNS. With our conceptual simplification of nested AD, these algorithms are now nearly a direct translation of Equations~\eqref{eq:predict} and~\eqref{eq:evidence}, with all of work done by the AD routines. We show the full 15-line Python implementation in the supplementary material. \ad{} and \adlns{} both have $O(K Y^{2.5})$ running time, where $Y = \sum_k y_k$.
\newcommand{\w}{0.24\textwidth}
\newcommand{\trim}{5pt}
\begin{figure}[t]
  \setlength\tabcolsep{0pt}
  \begin{tabular}{cc}
    \includegraphics[width=\w]{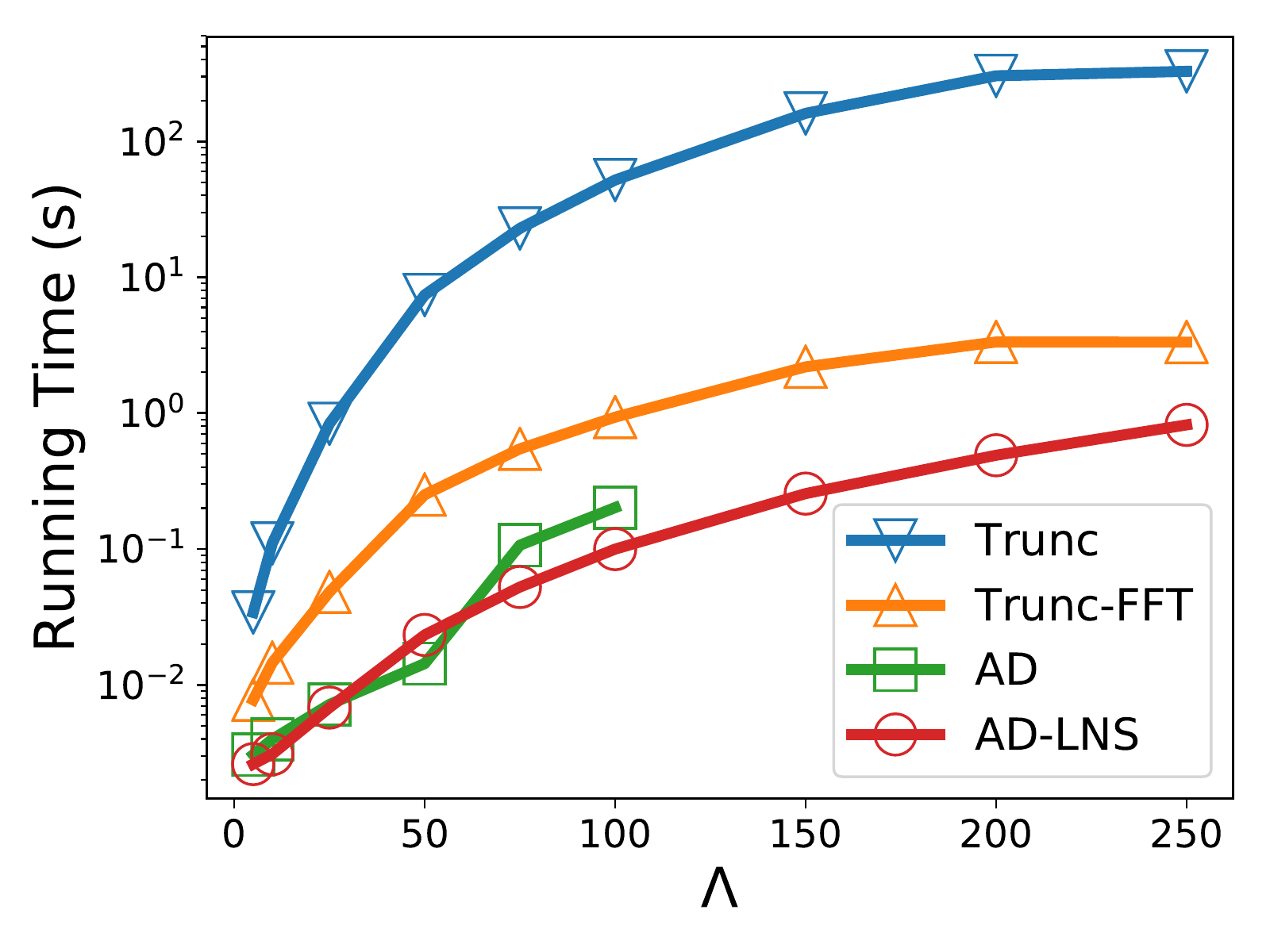} &
    \includegraphics[width=\w]{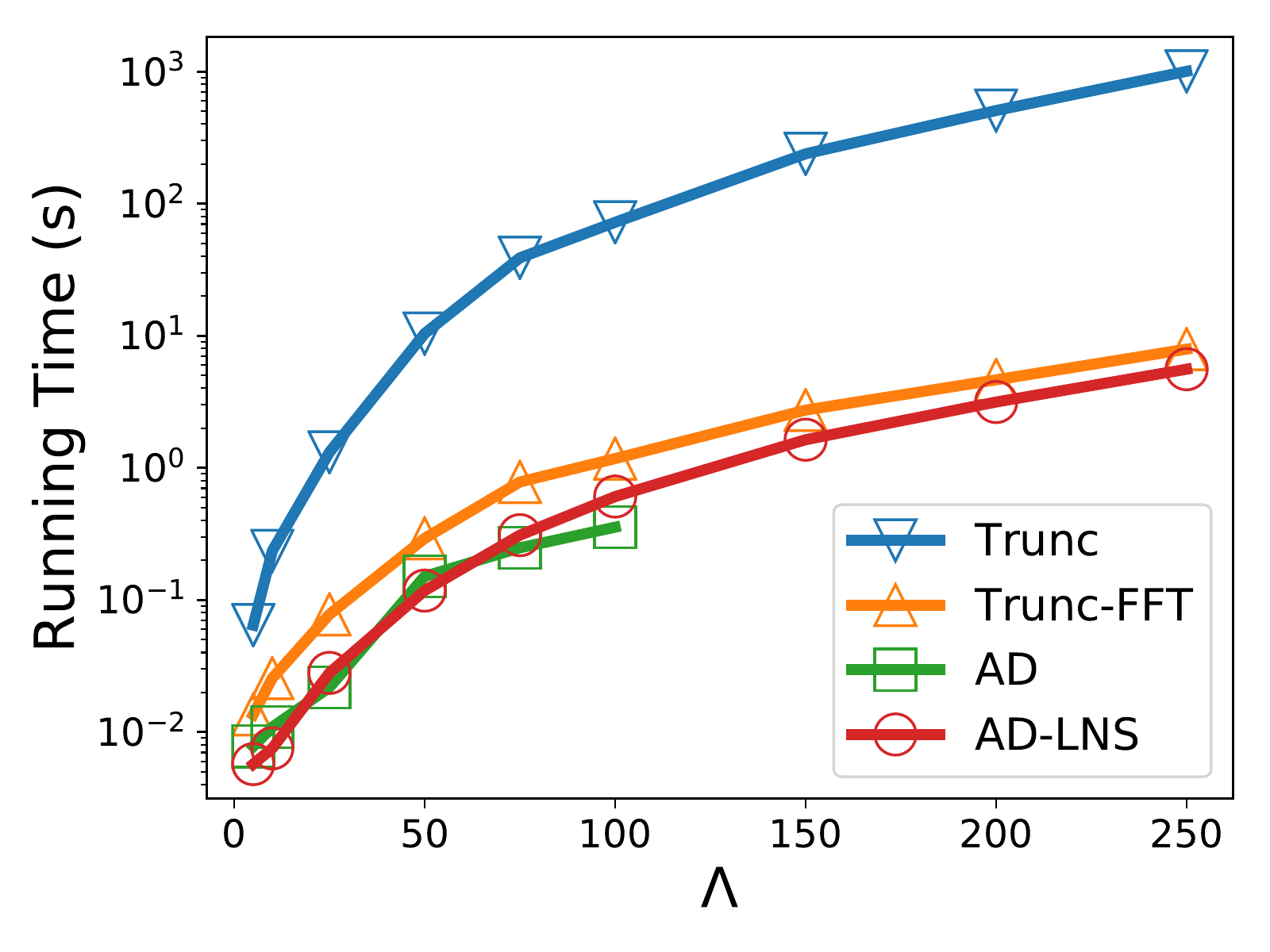} \\
    \includegraphics[width=\w]{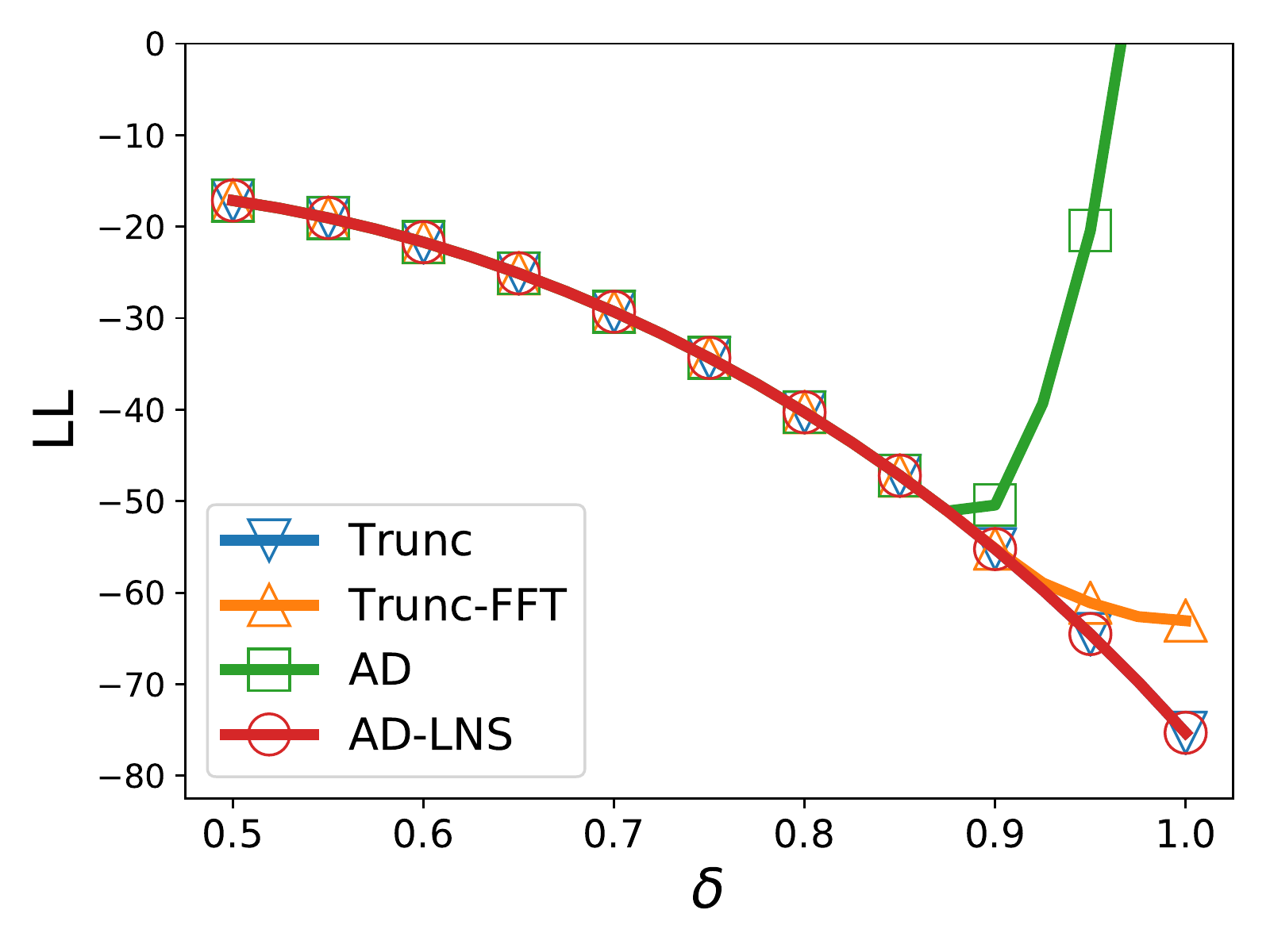} &
    \includegraphics[width=\w]{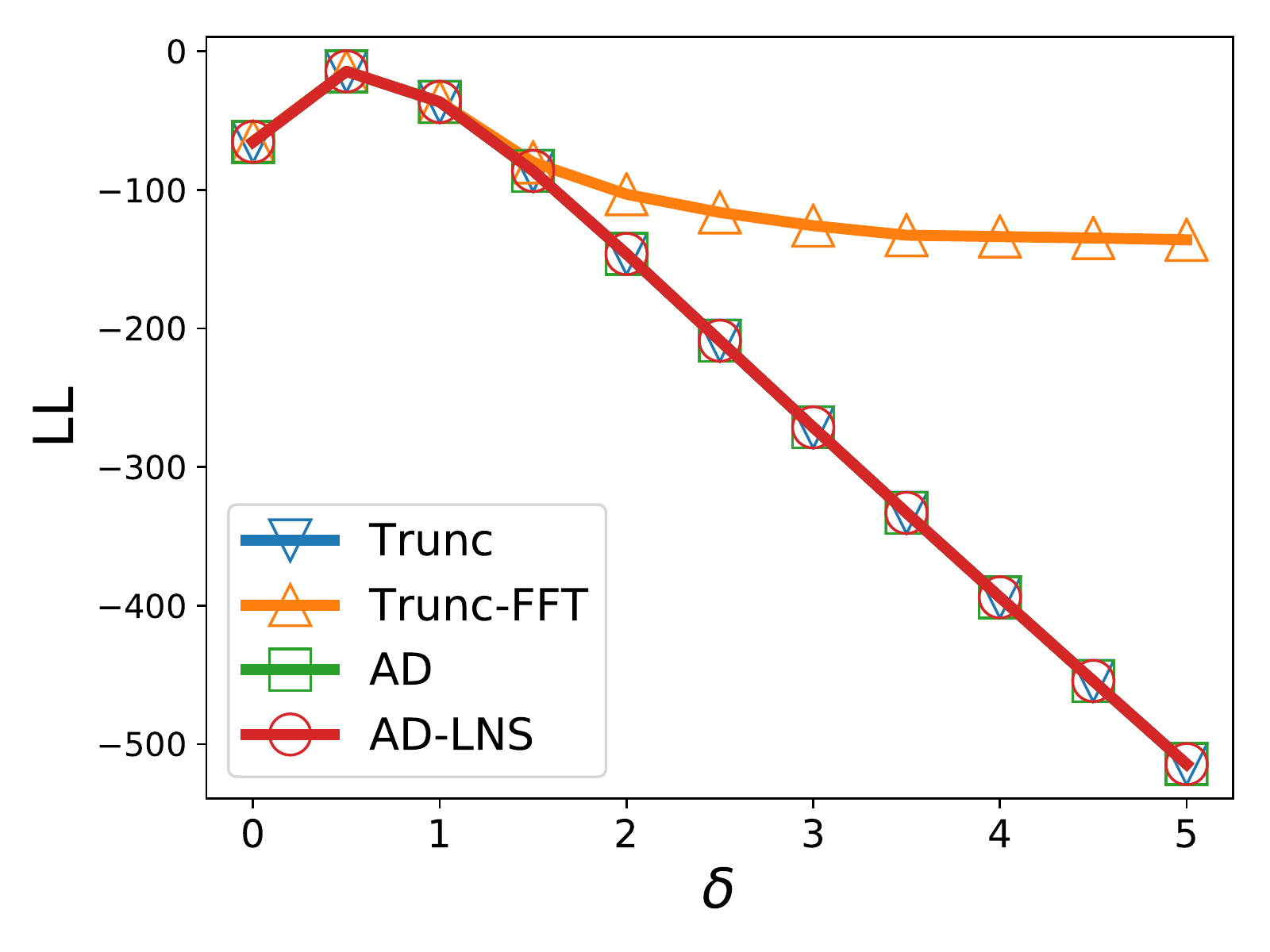} \\
    \\[-15pt]
    \small Bernoulli offspring &
    \small Poisson offspring \\
    %\small (c) Bernoulli &
    %\small (d) Poisson
  \end{tabular}
  \caption{\label{fig:inference}Accuracy, stability, and running time of inference. Top: running time vs. immigration parameter $\Lambda$ (population size). Bottom: Log-likelihood vs offspring parameter $\delta$. Left: Bernoulli offspring. Right: Poisson offspring.}
\end{figure}

We generated data from integer HMMs,
%with different offspring distributions and parameter values,
performed inference with each algorithm, and compared the resulting values and running times.
%computed log-likelihood values and running times.
In Figure~\ref{fig:inference}, top, we scaled the population size by generating data with immigration distribution $m_k \sim \text{Poisson}(\Lambda)$ for increasing $\Lambda$, and fixed offspring distributions $z_{k,i} \sim \text{Bernoulli}(0.5)$ or $z_{k,i} \sim \text{Poisson}(0.5)$.
%(i.e., roughly half of the individuals persist between time steps), and $\rho_k=0.5$.
Here, $\Lambda$ controls the immigration rate, and the expectation of all variables scales in proportion to $\Lambda$, as do the parameters $N$ and $Y$ controlling the running time of the algorithms. We ran inference using the true parameters.
%We used the four algorithms to compute the log-likelhood of the same parameters used to generate the data.
The plots show running time on a log scale. \ad{} and \adlns{} are the fastest algorithms, but \ad{} is unstable (due to overflow) and does not compute a result for $\Lambda > 100$. \truncfft{} is somewhat slower than the AD algorithms. % so for Bernoulli offspring than Poisson.
\adlns{} is especially fast for Bernoulli offspring because the Bernoulli PGF is linear, which leads to many zero power series coefficients, and LNS addition of zero is a very fast special case. The Poisson results should be considered the general case.

Figure~\ref{fig:inference}, bottom, compares the accuracy of different methods. Here, we generated data from a model with $\rho_k = 0.5$, immigration $m_k \sim \text{Poisson}(\lambda_k)$ for $\lambda_{1:5} = (12.5, 55, 105, 75, 20)$, and offspring distributions $z_{k,i} \sim \text{Bernoulli}(\delta)$ and $z_{k,i} \sim \text{Poisson}(\delta)$ with $\delta = 0.5$. We performed inference for \emph{different} values of $\delta$. This simulates the situation encountered during estimation when an optimizer queries the log-likelihood at an unlikely parameter setting; it is important that the methods are robust in this case. We see that as $\delta$ increases from the true value of $0.5$, \truncfft{} (in both models) and \ad{} (in Bernoulli) deviate from the correct answer. This is likely due to loss of numerical precision in floating point operations, either in the FFT algorithm or in power series algorithms.

Overall, \adlns{} is the only one of the four inference algorithms that is fast, stable (does not overflow or underflow), and accurate over a wide range of parameter settings.

%% \subsubsection{Number of Primitive Operations}
%% Because we need to implement basic arithmetic operations (+, --, *) in software for numbers in the logarithmic number system, these are much slower than floating point operations, which take only one or a few clock cycles. To compare the \emph{algorithms} directly, independently of the hardware, it's appropriate to compare them based on the total number of primitive arithmetic operations---either floating point or logarithmic number system depending on the algorithm. (This gives a view of the type of performance possible in the parallel universe where logarithmic number systems are widely adopted and implemented in hardware!)

%% Show results of one comparison using primitive operations instead of wall time.

\begin{figure}[t]
  \setlength\tabcolsep{0pt}
  \begin{tabular}{cc}
    \includegraphics[width=\w]{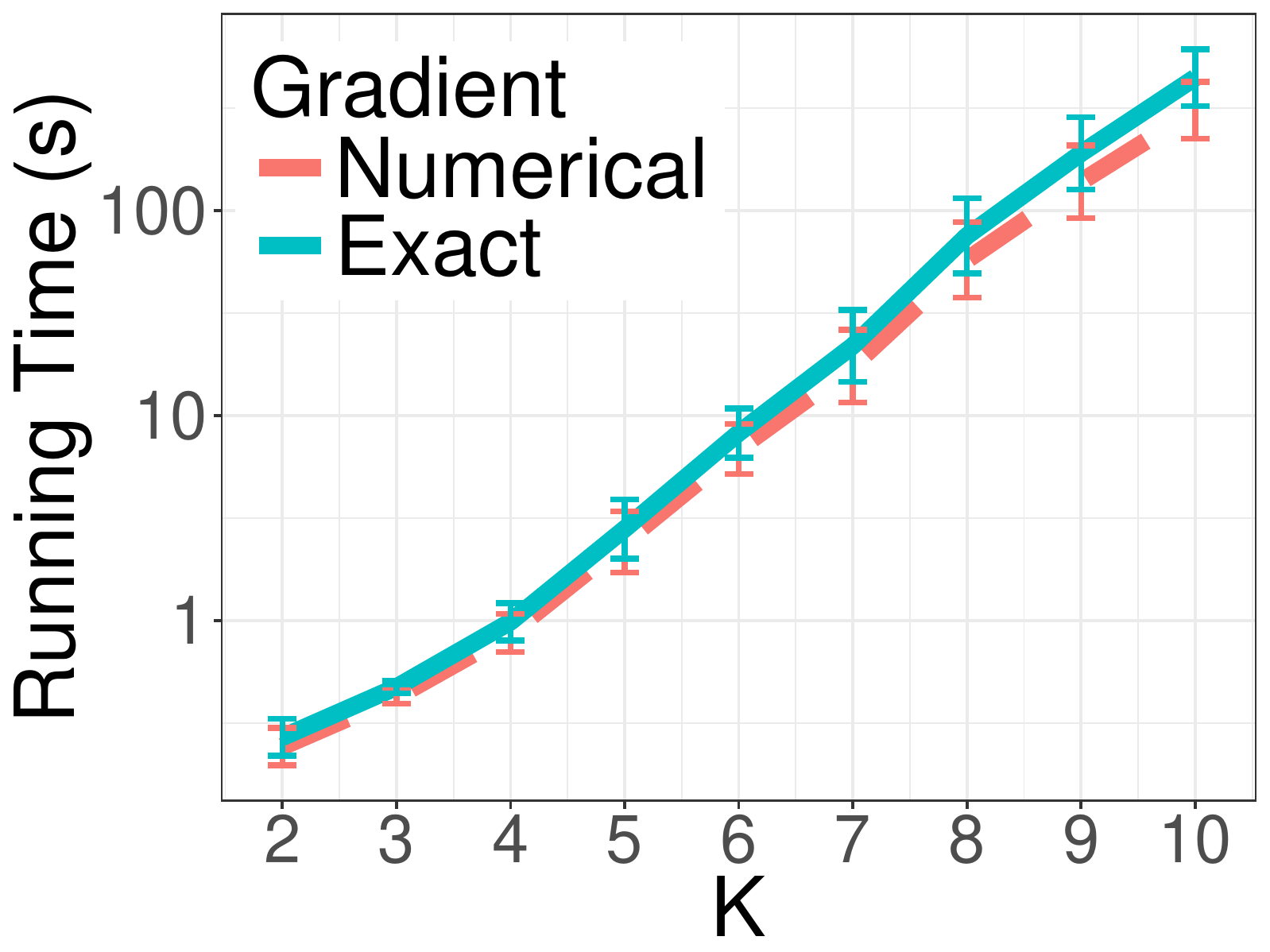} &
    \includegraphics[width=\w]{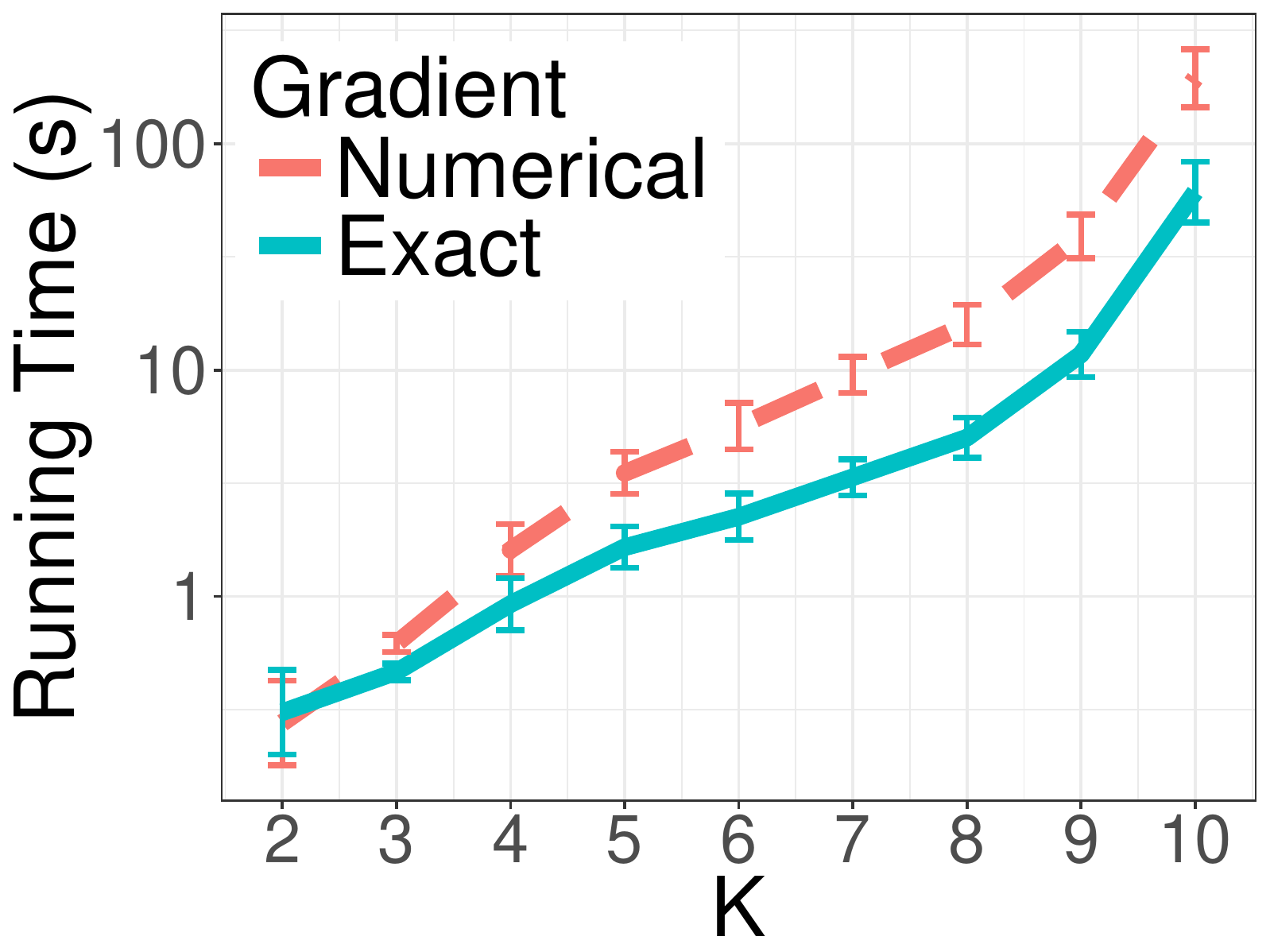} \\
    \small Single parameter & \small $K$ parameters\\[-6pt]
    %\\ (a) & (b) & (c) & (d)
  \end{tabular}
  \caption{\label{fig:mle-runtime}MLE running time vs number of time steps $K$. Left: with just one parameter, exact and numerical gradients are similar. Right: with $K$ parameters, optimization with exact gradients is much faster. See text for experiment details.  }
\end{figure}

\begin{figure}[t]
  \setlength\tabcolsep{0pt}
  \vspace{-8pt}
  \begin{tabular}{cc}
    \includegraphics[width=\w]{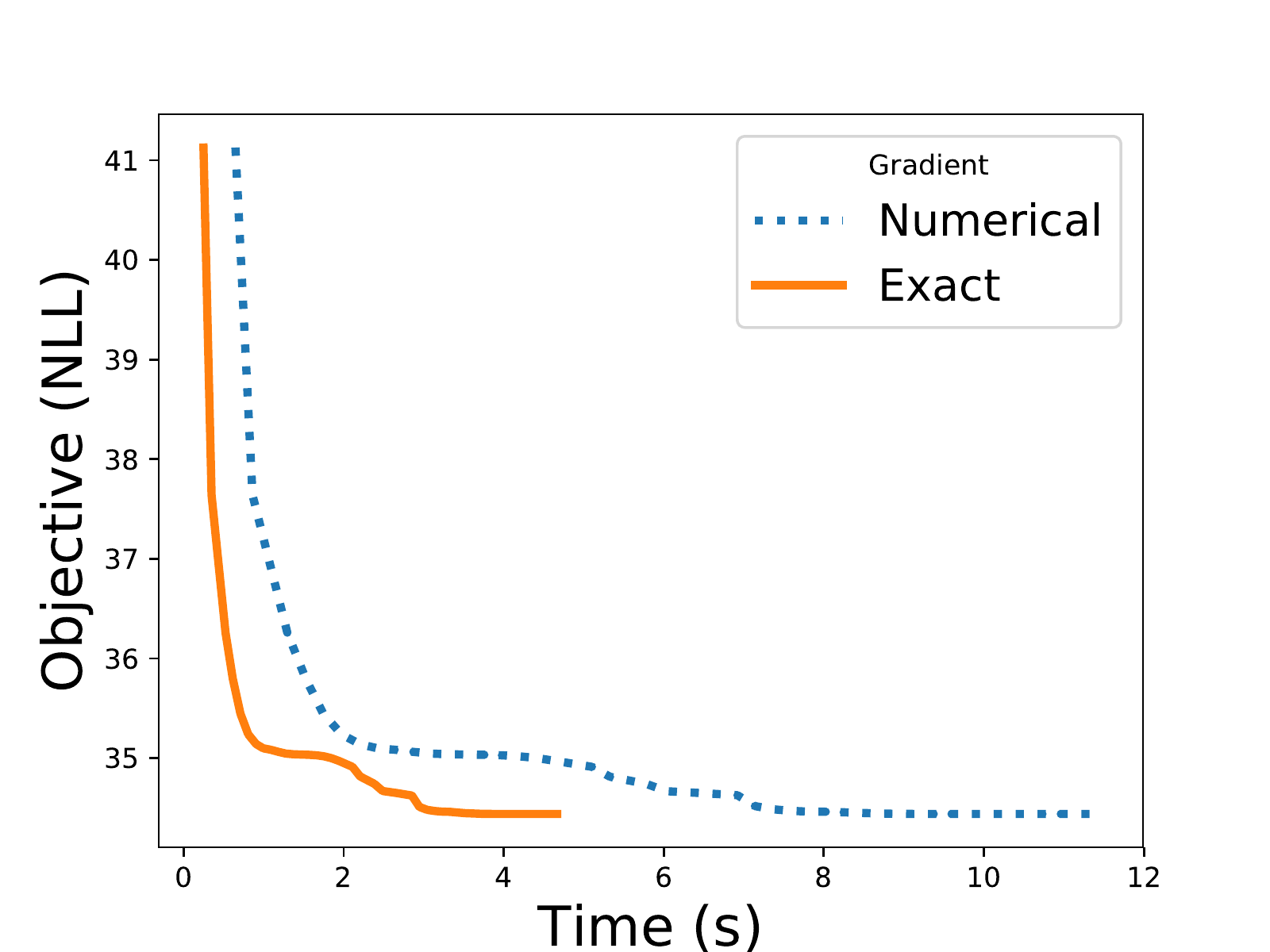} &
    \includegraphics[width=\w]{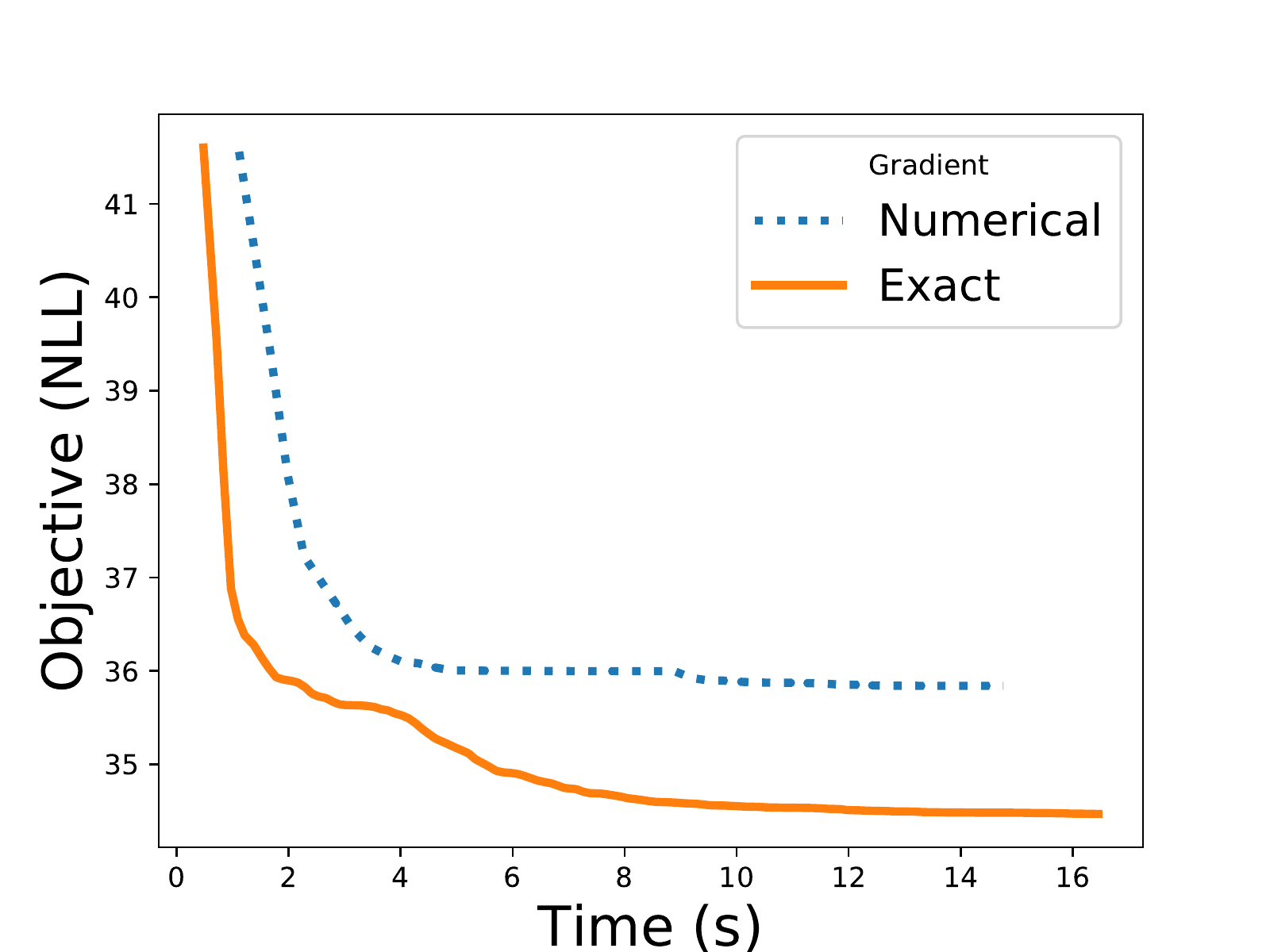} \\[-6pt]
    %\\ (a) & (b) & (c) & (d)
  \end{tabular}
  \caption{\label{fig:mle-trace}Objective trace of optimizer with different gradients on same problem. Left: a case where iterates are identical. Right: a case where exact gradients lead to a better solution.}
\end{figure}

\newcommand{\adback}{\textsc{AD-LNS-REV}}

\noindent
\textbf{Learning: Speed and Accuracy.}
We now evaluate the speed and accuracy of learning. We generate data from an integer HMM and use the L-BFGS~\cite{Liu1989} algorithm to recover parameters by maximizing the log-likelihood. Based on the inference results, we consider as a baseline the optimizer that uses \adlns{} to compute the likelihood and \emph{numerically} estimates gradients by finite differences.
%Due to either numerical instability, poor scalability, or inaccuracy, the other algorithms behave poorly within MLE routines for larger models.
The per-iteration cost to compute numerical gradients is the number of parameters times the cost to compute the log-likelihood (or twice this if central differences are used). For our approach, we used nested forward-over-reverse AD algorithm to compute exact gradients. In this case, the running time to compute the gradient is a constant factor greater than the time to compute the log-likelihood. We predict based on this that as the number of parameters increases, exact gradients will be faster.

Figure~\ref{fig:mle-runtime} shows the results. For each trial, we simulate 20 independent realizations of the integer HMM for up to $K=10$ time steps, with immigration distribution $m_k \sim \text{Poisson}(5)$ and $\rho = 0.6$. For the left plot, the offspring distribution $z_{k,i}\sim \text{Poisson}(1.2)$ is not time-varying, and we fit only the single parameter $\delta$. In this case, the cost of numerical gradient is only twice that of computing the objective, and the two methods behave similarly. Note that the running time is shown on a log scale. In the right plot, the offspring distribution $z_{k,i}\sim \text{Poisson}(\delta_k)$ is time-varying, with $\delta_k \sim \text{Exp}(1)$, and we fit each of these values for a total of $K$ parameters. In this case, we see that the exact gradient method is several times faster for $K \geq 4$. The running time of both methods increases with $K$ due to the increased cost of inference. We saw similar results across a wide range of models, and expect the benefits of using exact gradients to be even more significant for models with more parameters.

In most cases, the optimizer followed nearly the exact same sequence of parameter values regardless of the gradient computation, but was faster with exact gradients; in some cases the algorithm with exact gradients took more iterations and achieved a better objective value. See Figure~\ref{fig:mle-trace}.
%left, shows an example where the number of iterations and final value was the same for both methods, but the per-iteration cost was different. However, in models with too many parameters, the exact gradient method often took longer but converged to a better answer. Figure~\ref{fig:mle-trace}, right, shows an example. %The method with numerical gradients converges sooner, but to a higher objective value.

%This is likely due to very steep ``valleys'' in the negative log-likelihood surface corresponding to directions in parameter space that approximately preserve the log-likelihood, and numerical gradients not being precise enough to find descent directions through these valleys. In all cases like this, the two optimization methods behaved identically if we reduced the number of free parameters.

\section{Discussion}
We introduce new AD techniques for inference in integer HMMs. By cleanly incorporating nested derivatives into the AD computation model, we greatly simplify the algorithms for inference using PGFs. By implementing forward-mode AD operations using the logarithmic number system and using fast power series algorithms, we achieve fast, accurate, and stable algorithms for very high-order derivatives. Our new techniques for nested forward-over-reverse AD allow us to compute exact gradients in integer HMMs for the first time, which leads to significantly faster, and sometimes more accurate, learning procedures.

\textbf{Related work.} An alternate approach to compute nested derivatives is to repeatedly apply AD at each level of nesting starting with the innermost, for example, using source transformation. \citet{Pearlmutter2007} and \citet{Siskind2008} introduced a tagging mechanism for forward AD to properly handle nesting and avoid ``perturbation confusion''~\cite{siskind2005perturbation}; this idea is impemented in functional AD tools such as DiffSharp~\cite{Baydin2015}. However, since each application of AD increases running time by a constant factor, and the innermost function is ``transformed'' each time (either through source transformation or by adding a new tagged perturbation),
the running time of these approaches is exponential in the number of levels of nesting, while our running time is polynomial. But note that these methods can handle more general functions than ours (i.e., multivariate, not subject to Assumption~1). An interesting avenue of future work is to extend our approach to multivariate nested derivatives.

 %by introducing formal $\varepsilon$ variables to track perturbations corresponding to different nesting levels. 
%address nesting in a functional framework; these methods can handle more general functions than ours (i.e., multivariate, not subject to Assumption~1) but are also exponential in the number of nesting levels due to 

\section*{Acknowledgments}
This material is based upon work supported by the National
Science Foundation under Grants No. 1617533 and 1522054.

\eat{
\section{Comparison to Existing AD Approaches}
\red{Discuss other AD methods that can handle nesting, and exponential time complexity? Do this last.}
\red{Note limitation: nesting of \emph{univariate} functions. Use figure}
* Source-to-source from inside out: obviously exponential. Innermost code gets doubled $K$ times.
* Tower method of Siskind and Pearlmutter. Exponential, because a new $\epsilon$ is instantiated at each level of nesting, and coefficients are doubled. (Run this? Doubtful.)
* Express as mixed partial derivative of a multivariate function. Multivariate higher-order forward AD. Also exponential.
}

\bibliography{pgf-backprop}
\bibliographystyle{icml2018}

\clearpage

\appendix

\section{Definition of Partial Computation}
Formally, we can write $f_{i \to \ell}$, the partial computation from $v_i$ to $v_\ell$, via the recurrence:
\begin{align*}
f_{i\to\ell}(v_{0:i}) &= \varphi_\ell \big(u_{ij}(v_{0:i})\big)_{j \in A_\ell}, \\
u_{ij}(v_{0:i}) &=
\begin{cases}
  v_{j}          & \text{if } j \leq i \\
  f_{i\to{}j}(v_{0:i}) & \text{if } j > i
\end{cases}
\end{align*}

Here, $u_{ij}(v_{0:i})$ will equal $v_j$ for all $j$. For $j \leq i$, the value is obtained directly from the inputs to $f_{i \to \ell}$. For $j > i$, the value of $u_{ij}$ is computed according to the partial computation from $i$ to $j$.

\section{Proof of Proposition 3}

\begin{proof}
%%   The Taylor expansion of $f_{0 \to k}$ about $x$ is:
%%   \begin{align}
%%   \label{eq:fk-taylor}
%%   f_{0\to k}(x + \epsilon) &= v_k + \underbrace{\sum_{i=1}^\infty \frac{f^{(i)}_{0 \to k}(x)}{i!} \epsilon^i}_{R(\epsilon)}.
%% \end{align}
%% The Taylor series of $\varphi_j$ about $v_k$ is: 
%% \begin{equation}
%%   \label{eq:varphi-taylor}
%%   \varphi_j(v_k + \tau) = \varphi_j(v_k) + \underbrace{\sum_{i=1}^\infty \frac{\varphi_j^{(i)}(v_k)}{i!} \tau^i}_{Q(\tau)}.
%% \end{equation}
We write $f_{0 \to j}$ as a function of $\epsilon$ using the two given Taylor expansions:
\begin{align*}
  f_{0 \to j}(x + \epsilon)
  &= \varphi_j\big(f_{0 \to k}(x + \epsilon)\big), \\
  &= \varphi_j\Big(v_k + \big(f_{0 \to k}(x + \epsilon)-v_k\big)\Big), \\
  &= v_j + Q\Big(f_{0 \to k}(x+\epsilon) - v_k \Big), \\
  &= v_j + Q\big(R(\epsilon)\big),
\end{align*}
where we used the Taylor expansion of $\varphi_j$ in Line 3 and the Taylor expansion of $f_{0 \to k}$ in Line 4. The last expression is a power series in $\epsilon$, and, since $f_{0 \to j}$ is analytic (the composition of two analytic functions is analytic), it is necessarily the Taylor series expansion of $f_{0 \to j}$ about $x$.
\end{proof}

\section{Truncated Forward Algorithm}
The truncated forward algorithm is the following variant of the forward algorithm for discrete HMMs, where $N$ is an upper bound placed on the population size:
\begin{enumerate}
\item Set $\alpha_0(0) = 1$ and $\alpha_0(n) = 0$ for $n=1$ to $N$.
\item For $k=1$ to $K$
  \begin{enumerate}
  \item Compute the transition matrix $P_k$, where $P_k(n,n') = \Pr(n_k = n' \mid n_{k-1} = n)$ for all $n, n' \in \{0, \ldots, N\}$ (details below)
    \item For $n' = 0$ to $N$, set
      \[
      \alpha_k(n') = p(y_k \mid n') \sum_{n = 0}^{N} \alpha_{k-1}(n) P_k(n, n')
      \]
  \end{enumerate}
\item The likelihood is $\sum_{n=0}^N \alpha_K(n)$
\end{enumerate}

Step 2(b) takes $O(N^2)$ time. Step 2(a) may take $O(N^3)$ or $O(N^2 \log N)$ time, depending on how it is implemented. Observe that $n_k$ is the sum of two random variables:
\[
n_k = z_k + m_k, \qquad  z_k := \sum_{i=0}^{n_{k-1}} z_{k,i}
\]
and we must reason about their convolution to construct the transition probabilities. Specifically, the $n$th row of $P_k$ is the convolution of the first $N$ values of the distribution $p(z_k \mid n_{k-1} = n)$ and the first $N$ values of $p(m_k)$:
\[
P_k(n,n') = \sum_{z=0}^N \Pr(z_k = z \mid n_{k-1} = n) \Pr(m_k = n' - z)
\]
Let us assume we can compute the first $N$ values of each distribution in $O(N)$ time. Then the time to compute each row of $P_k$ is $O(N^2)$ if we use the direct convolution formula above, but $O(N \log N)$ if we use the fast Fourier transform (FFT) for convolution, making the overall procedure either $O(K N^3)$ or $O(K N^2 \log N)$. While the FFT is superior in terms of running time, it is inaccurate in many cases (see Section~\ref{sec:experiments}).

\begin{figure*}[t!]
\begin{lstlisting}[language=python]
def A(s, k):        
    if k < 0: return 1.0

    # This allows constant to be constructed in log space
    const = GDual.const( y[k] * np.log(rho[k]) - gammaln(y[k] + 1), as_log=True )

    return (s**y[k]) * const * \
        diff( lambda u: Gamma(u, k), s*(1 - rho[k]), y[k] )

def Gamma(u, k):
    F = lambda u:   offspring_pgf( u, theta_offspring[k-1] )
    G = lambda u: immigration_pgf( u, theta_immigration[k] )
    return A(F(u), k-1) * G(u)

log_likelihood = log( A(1.0, K-1) )
\end{lstlisting}
\caption{Code for AD based forward algorithm.}
\end{figure*}

\end{document}